\documentclass[10pt,twocolumn,letterpaper]{article}

\usepackage[pagenumbers]{cvpr} 

\usepackage{url}
\usepackage{booktabs}
\usepackage{multirow}
\usepackage{colortbl}
\usepackage{graphicx}
\usepackage{float}
\usepackage{xcolor}

\usepackage{graphicx}
\usepackage{amsmath}
\usepackage{amssymb}
\usepackage{booktabs}
\usepackage{soul}
\usepackage{algorithm}
\usepackage{algorithmic}

\usepackage[pagebackref,breaklinks,colorlinks]{hyperref}

\usepackage[capitalize]{cleveref}
\crefname{section}{Sec.}{Secs.}
\Crefname{section}{Section}{Sections}
\Crefname{table}{Table}{Tables}
\crefname{table}{Tab.}{Tabs.}

\def\cvprPaperID{9500} 
\def\confName{CVPR}
\def\confYear{2023}

\begin{document}

\title{Towards Domain Generalization for \\ Multi-view 3D Object Detection in Bird-Eye-View}

\author{Shuo Wang\textsuperscript{1*} \quad Xinhai Zhao\textsuperscript{2*} \quad Hai-Ming Xu\textsuperscript{3} \quad Zehui Chen\textsuperscript{1} \quad Dameng Yu\textsuperscript{2} \\
\quad Jiahao Chang \textsuperscript{1} 
\quad Zhen Yang\textsuperscript{2} 
\quad Feng Zhao\textsuperscript{1$\dagger$} \\
\textsuperscript{1}University of Science and Technology of China\\
\textsuperscript{2}Huawei Noah’s Ark Lab     \\
\textsuperscript{3}University of Adelaide\\
}
\maketitle

\setlength{\skip\footins}{1.0cm}
\renewcommand{\thefootnote}{\fnsymbol{footnote}} 
\footnotetext[1]{Shuo Wang and Xinhai Zhao contributed equally. This work was done when Shuo Wang was an intern at Huawei Noah's Ark Lab.} 
\footnotetext[2]{Corresponding author.} 


\begin{abstract}
    Multi-view 3D object detection (MV3D-Det) in Bird-Eye-View (BEV) has drawn extensive attention due to its low cost and high efficiency. Although new algorithms for camera-only 3D object detection have been continuously proposed, most of them may risk drastic performance degradation when the domain of input images differs from that of training. In this paper, we first analyze the causes of the domain gap for the MV3D-Det task. Based on the covariate shift assumption, we find that the gap mainly attributes to the feature distribution of BEV, which is determined by the quality of both depth estimation and 2D image's feature representation. To acquire a robust depth prediction, we propose to decouple the depth estimation from the intrinsic parameters of the camera (i.e. the focal length) through converting the prediction of metric depth to that of scale-invariant depth and perform dynamic perspective augmentation to increase the diversity of the extrinsic parameters (i.e. the camera poses) by utilizing homography. Moreover, we modify the focal length values to create multiple pseudo-domains and construct an adversarial training loss to encourage the feature representation to be more domain-agnostic. Without bells and whistles, our approach, namely DG-BEV, successfully alleviates the performance drop on the unseen target domain without impairing the accuracy of the source domain. Extensive experiments on various public datasets, including Waymo, nuScenes, and Lyft, demonstrate the generalization and effectiveness of our approach. To the best of our knowledge, this is the first systematic study to explore a domain generalization method for MV3D-Det.
\end{abstract}
    

\section{Introduction}
\label{sec:intro}
3D object detection, aiming at localizing objects in the 3D space, is critical for various applications such as autonomous driving \cite{wang2021multi,chen2022autoalign}, robotic navigation~\cite{antonello2017fast}, and virtual reality~\cite{schuemie2001research}, \etc. 
Despite the remarkable progress of LiDAR-based methods~\cite{lang2019pointpillars,qi2017pointnet,shi2020pv}, camera-based 3D object detection in Bird-Eye-View (BEV)~\cite{huang2021bevdet,li2022bevformer,li2022bevdepth} has drawn increasing attention in recent years due to its rich semantic information and low cost for deployment. 

\begin{figure}[tbp]
\centering
\begin{minipage}[t]{0.22\textwidth}
\includegraphics[width=4cm]{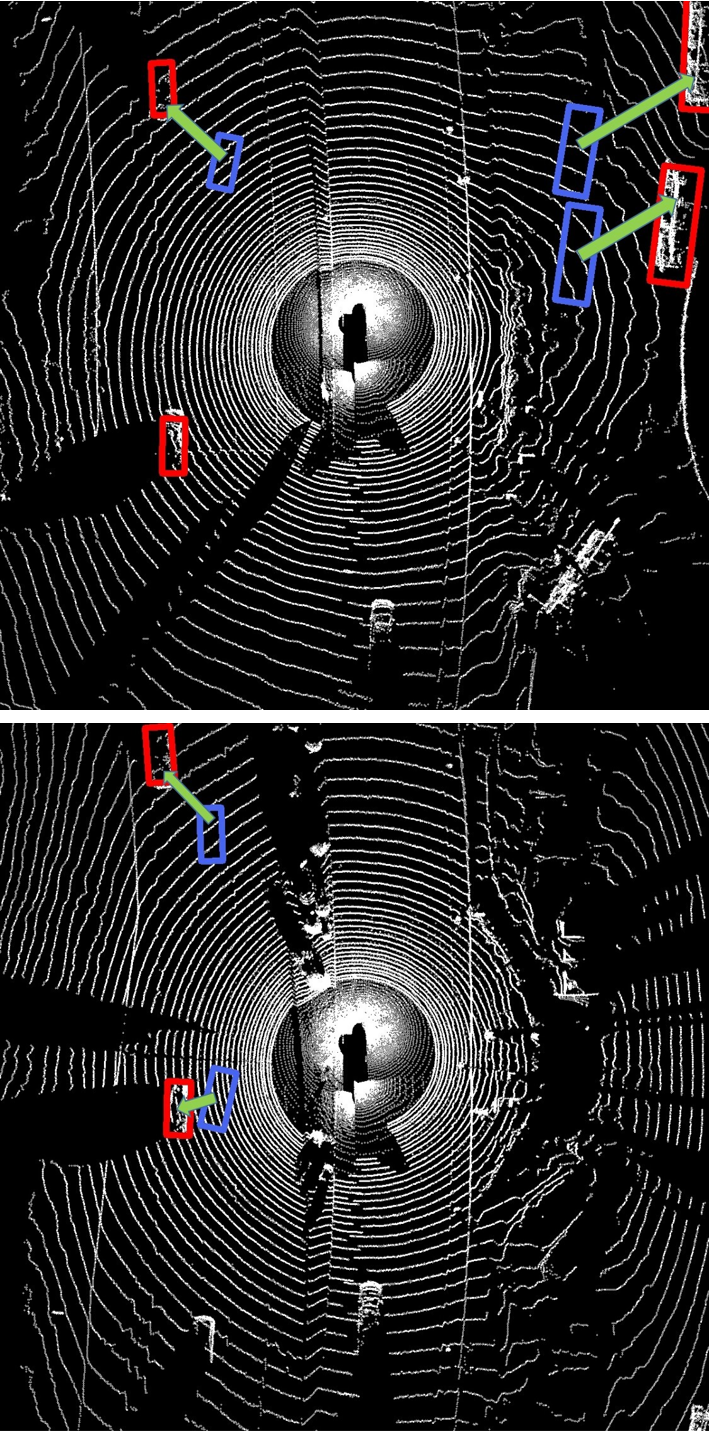}
\centerline{(a) Baseline}\medskip
\end{minipage}
\hfill
\begin{minipage}[t]{0.22\textwidth}
\includegraphics[width=4cm]{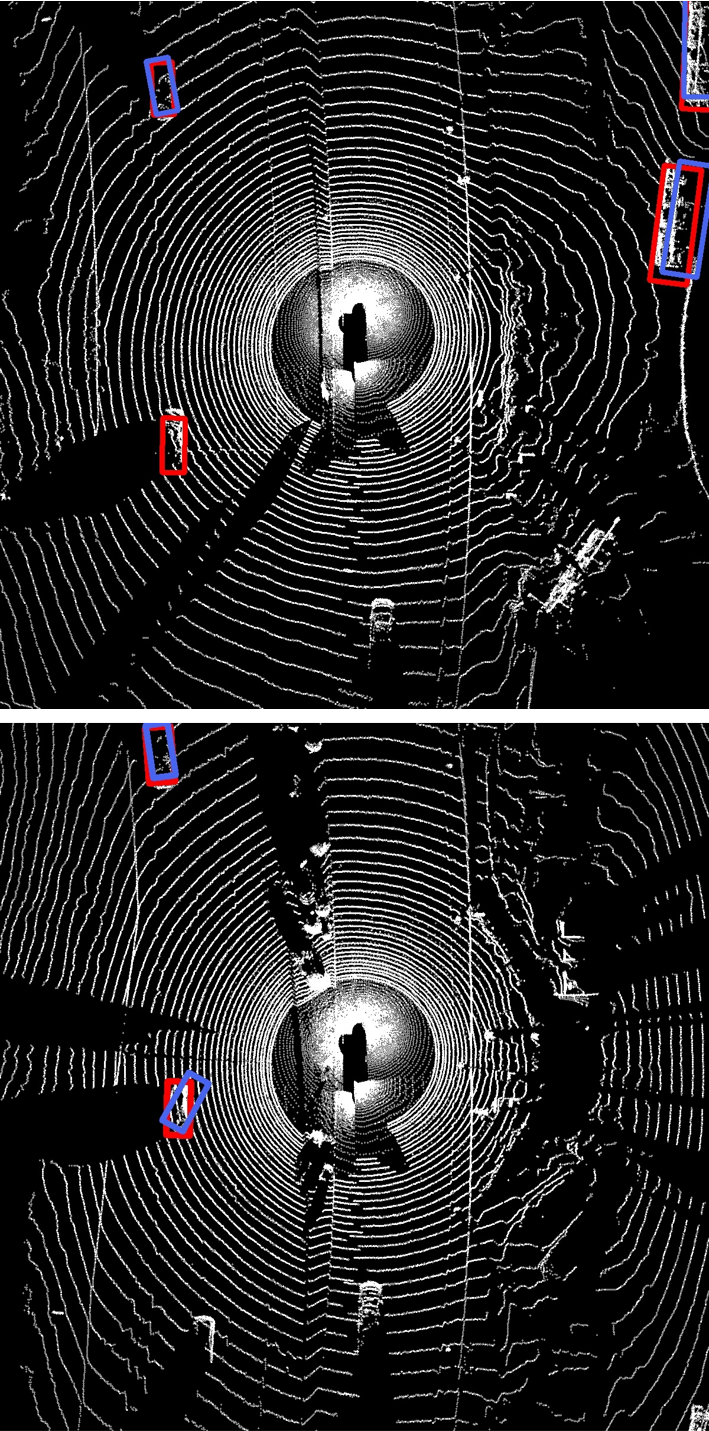}
\centerline{(b) DG-BEV}
\end{minipage}
\caption{Qualitative comparisons between BEVDepth and the proposed DG-BEV. The red and blue
bounding boxes represent ground truth and detected results on the target domain respectively. Depth-shift is shown in green arrows.
Our approach can detect correct 3D results on unknown domains.}
 \label{fig:intro}
\end{figure}

However, most of the detectors assume that the training and testing data are obtained in the same domain which may be hardly guaranteed in realistic scenarios. Thus, tremendous performance degradation will appear when the domain of the input image shifts.
For example, nuScenes~\cite{caesar2020nuscenes} and Waymo~\cite{sun2020scalability} are two popular benchmarks for 3D object detection and their data collection devices are not identical, \ie, both of the intrinsic and extrinsic parameters are different. Empirical results presented in \cref{fig:intro} show that detectors trained on nuScenes have location bias when predicting objects on the Waymo dataset.

\begin{figure}[!t]
  \centering
   \includegraphics[width=0.8\linewidth]{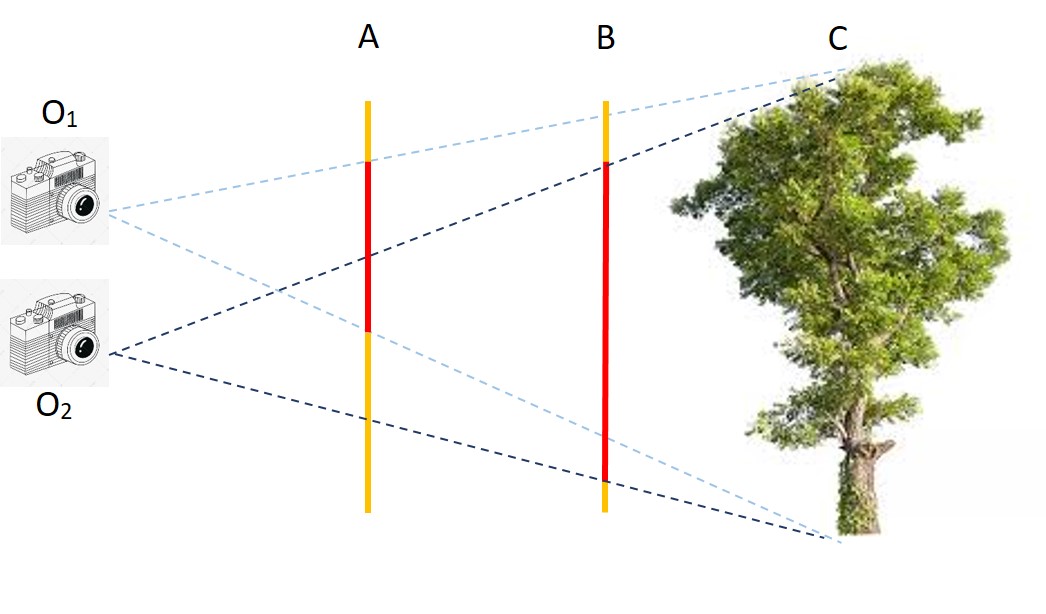}
   \vspace{-1em}
   \caption{Illustration of the difficulty in estimating depth based on cameras with different focal length. $O_1$ and $O_2$ are the optical centers of two cameras and C is the object being photographed. A and B denote the imaging planes of the two cameras respectively and the red parts show the size of the same object in their corresponding image planes.}
   \label{fig:error}
\end{figure}

 Domain Generalization (DG)~\cite{muandet2013domain,li2018domain,dou2019domain}, aiming to learn a model that generalizes well on unseen target domains, can be a plausible solution to alleviate the bias mentioned above. In the literature, DG has been widely explored for 2D vision tasks, \eg, image recognition~\cite{keysers2007deformation,dosovitskiy2020image}, object detection~\cite{ren2015faster,zhu2020deformable}, and semantic segmentation \cite{yu2018bisenet,noh2015learning}. However, most of these works are designed for the case where there are multiple source domains available which are obviously infeasible due to the diversity of the real world in autonomous driving scenarios.
Alternatively, one recent work~\cite{wu2022single} proposed to study the single-domain generalization for LiDAR-based detection. However, it is not tractable to directly adapt this method to solve the camera-based detection task due to the fundamental differences between the characteristics of points and images. Therefore, developing a general domain generalization framework for MV3D-Det is still highly desirable.

In this paper, we theoretically analyze the causes of the domain gap for MV3D-Det. Based on the covariate shift assumption~\cite{chen2018domain}, we find that such a gap mainly attributes to the feature distribution of BEV, which is determined by the depth estimation and 2D image feature jointly. 
Based on this, we propose DG-BEV, a domain generalization method for MV3D-Det in BEV. Specifically,
we first conduct a thorough analysis of why the estimated depth becomes inaccurate when the domain shifts and find the key factor lies in 
that intrinsic parameters of cameras used in various domains are hardly guaranteed to be identical (please refer to~\cref{fig:error} for a better understanding). To alleviate this issue, we propose to decouple the depth estimation from the intrinsic parameters by converting the prediction of metric depth to that of scale-invariant depth. 
On the other hand, extrinsic parameters of cameras (\eg camera poses) also play an important role in camera-based depth estimation, which is often ignored in previous works. Instead, we introduce homography learning to dynamically augment the image perspectives by simultaneously adjusting the imagery data and the camera pose.

Moreover, since domain-agnostic feature representations are favored for better generalization, we propose to build up multiple pseudo-domains by modifying the focal length values of camera intrinsic parameters in the source domain and construct an adversarial training loss to further enhance the quality of feature representations.
In summary, the main contributions of this paper are:
\begin{itemize}
\item[$\bullet$] We present a theoretical analysis on the causes of the domain gap in MV3D-Det. Based on the covariate shift assumption, we find the gap lies in the feature distribution of BEV, which is determined by the depth estimation and 2D image feature jointly.
\item[$\bullet$] We propose DG-BEV, a domain generalization method to alleviate the domain gap from both of the two perspectives mentioned above.

\item[$\bullet$] Extensive experiments on various public datasets, including Waymo, nuScenes, and Lyft, demonstrate the generalization and effectiveness of our approach. 

\item[$\bullet$] To the best of our knowledge, this is the first systematic study to explore a domain generalization method for multi-view 3D object detectors. 
\end{itemize}

\section{Related Works}

\subsection{Vision-based 3D object detection}
Vision-based 3D object detection~\cite{ma2022vision} is gaining more and more attention from researchers due to rich semantic information and low cost for deployment. In the last few years, many efforts have been made on predicting objects directly from a single image. For example, inspired by FCOS~\cite{tian2019fcos}, FCOS3D~\cite{wang2021fcos3d} extends this paradigm to 3D object detection and achieves great performance. Since single view-based prediction does not integrate information from multiple cameras well, there is a growing interest in MV3D-Det.
LSS~\cite{philion2020lift} is the first to explore the mapping of multi-view features to BEV space. Based on LSS, BEVDet~\cite{huang2021bevdet} enables this paradigm to perform competitively. BEVDepth~\cite{li2022bevdepth} regard LiDAR as supervisory information for depth and enhance the model’s ability of depth perception. DETR3D~\cite{wang2022detr3d} integrates information from multiple perspectives in an attention pattern and GraphDETR3D~\cite{chen2022graph} improves performance further by utilizing graph neural networks. Moreover, PETR ~\cite{liu2022petr}
proposes 3D position-aware encoding, which greatly improves the performance of DET3D.

\subsection{Domain Adaption and Domain Generalization}
Domain adaptation (DA) aims to improve models’ performance on a known target domain. Many approaches have been designed for 2D detection. Particularly, ~\cite{chen2018domain} proposed to align both feature-level and instance-level distributions through an adversarial mechanism~\cite{ganin2015unsupervised}. Subsequent work~\cite{xu2020exploring,he2020domain,zhao2020collaborative,acuna2021towards} expands on this foundation. Since sometimes we can not get access to the target domain, some studies have started to focus on domain generalization (DG)~\cite{muandet2013domain,li2018domain,dou2019domain,facil2019cam}, which targets generalizing a model trained on source domains to many unseen target domains. 

However, the aforementioned methods based on 2D detection mainly focus on handling lighting, color, and texture variations. Obviously, they can not be directly applied to 3D detection, the focus of which is to accurately estimate the spatial information of objects. Thus, some domain adaption methods specific to 3D perception have been explored. CAM-Convs~\cite{facil2019cam} is a new type of convolution that improves the generalization capabilities of depth prediction networks considerably. For LiDAR-based detection~\cite{wu2022single}, differences in data structures and network architectures make it impossible to apply to multi-view 3D detection. STMono3D~\cite{li2022unsupervised} only explores single-view 3D detection instead of multi-view.

\begin{figure*}[h]
  \centering
   \includegraphics[width=1.0\linewidth]{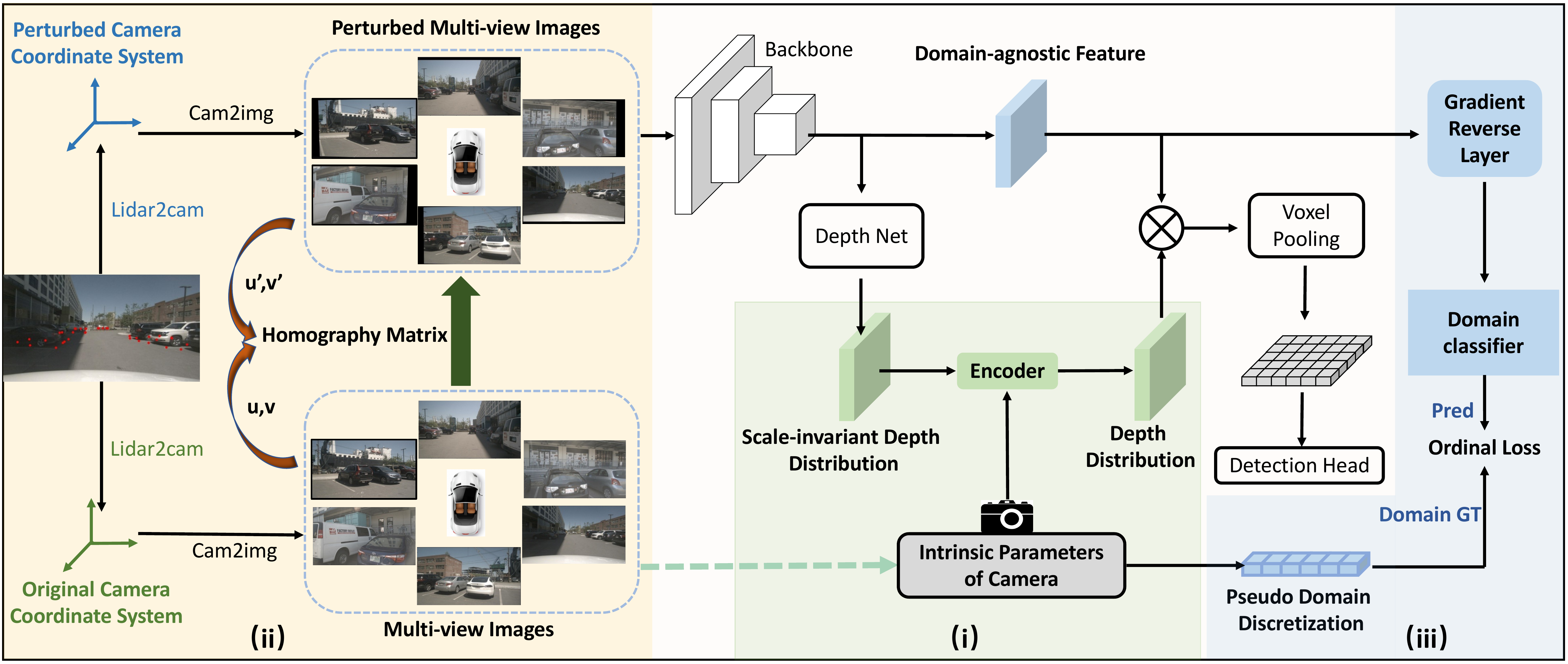}
   \vspace{-1em}
   \caption{The overall framework of our approach DG-BEV. Building on top of BEVDepth, we propose three efficient strategies to 
    improve the domain generalization ability:
   (i) intrisics-decoupled depth estimation in \cref{sec:3.3.1} (ii) dynamic perspective augmentation in \cref{subsec:pose_aug}, and (iii) domain-invariant feature learning in \cref{subsec:domain_inv_feat}.
}
   \label{fig:network}
\end{figure*}

\section{Method}

\subsection{Problem Definition}
Under the domain generalization setting, we can access the labeled images from the source domain $D_S=\{x_s^i, y_s^i, K_s^i, E_s^i\}_{i=1}^{N_S}$ but the target domain $D_T=\{x_t^i, y_t^i, K_t^i, E_t^i\}_{i=1}^{N_T}$ is not available, of which $N_s$ and $N_t$ are the numbers of samples from the source and target domains, respectively. Each 2D image $x^i$ comes with the camera intrinsic parameter $K^i$ and the extrinsic parameter $E^i$. $K^i$ is responsible for projecting the points in 3D space to the 2D image plane and $E^i$ indicates the camera pose, which is composed of yaw, pitch, and roll. Label $y^i$ consists of object class $k$, location $(c_x, c_y, c_z)$, size in each dimension $(d_x, d_y, d_z)$, and orientation $\theta$. We aim to train models with $D_S$ and achieve as good results as possible when inferring in any other target domain $D_T$. At the same time, the process described above will not impair the accuracy of the source domain.

\subsection{A Probabilistic View of the Domain Gap}
MV3D-Det in Bird-Eye-View (BEV) can be viewed as a component of two parts, one is a mapping that projects a 2D image into the feature map of 3D space (\ie BEV), and the other one is to learn the posterior $P(Y|X, K, E)$, where $Y$ is the ground truth consisting of category, location, dimension and orientation, $X$ is the image representation, $K$ is the intrinsic parameter and $E$ is the extrinsic parameter.
Let $P_S(Y, X, K, E)$ and $P_T(Y, X, K, E)$ represent the joint distribution of training samples in the source domain and the target domain, respectively. When there exists no domain shift in theory, it means that $P_S(Y, X, K, E)=P_T(Y, X, K, E)$.
According to Bayes's Formula, we decompose the joint distribution as:
\begin{equation}
  P(Y, X, K, E)=P(Y|X, K, E)P(X, K, E).
  \label{eq:important}
\end{equation}
Similar to~\cite{chen2018domain}, we make the covariate shift assumption for $P(Y|X, K, E)$, \ie, different domains naturally have the same conditional probability, and the domain distribution shift results from the inconsistent marginal distribution $P(X, K, E)$.
In MV3D-Det, $P(X, K, E)$ indicates the feature distribution of 2D images projected into 3D space, which is determined by the depth estimation and 2D image feature jointly. Hence, we try to improve the existing domain shift from the above two aspects.

\subsection{DG-BEV}

In this section, we introduce our domain generation framework for multi-view 3D object detection, DG-BEV. Building on top of BEVDepth, we designate three simple approaches: (i) intrinsic-decoupled depth prediction, (ii) dynamic perspective augmentation, and (iii) domain-agnostic feature learning. \cref{fig:network} illustrates the overall framework of our approach.

\subsubsection{Intrinsics-Decoupled Depth Prediction\label{sec:3.3.1}}
As shown in~\cref{fig:error}, when two cameras with various intrinsic parameters (\ie focal lengths) shoot the same object at the same distance, the imaging size of the object, which is determined by the intrinsic parameters, can be quite different. If a model is only optimized on the dataset collected from a specific camera, it can be difficult for the model to predict an identical depth for the object pictured from another camera, and thus it is the cause of inaccurate depth prediction when domain shifts, similar in~\cite{li2022unsupervised}. 
Furthermore, we empirically find that the estimated depth has been entangled with the intrinsic parameters of the camera, and results in non-compliance with the intuition of ``Everything looks small in the distance and big on the contrary". Hence, we attempt to decouple the estimated depth from the intrinsic parameters. 

Random scaling of an image is one of the widely used augmentation methods. When an image is randomly scaled, the intrinsic parameter can be denoted as 

\begin{equation}
  K =    {
        \left[ \begin{array}{ccc}
        r_x & r_y & 1
        \end{array} 
        \right ]}
        {
        \left[ \begin{array}{ccc}
        f_x & 0 & p_x\\
        0 & f_y & p_y\\
        0 & 0 & 1
        \end{array} 
        \right ]},
  \label{eq:1}
\end{equation}
where $ r_x$ and $r_y$ are resize rates, $f$ and $p$ are the focal length and optical center, x and y indicates image coordinate axes, respectively. 

According to~\cref{eq:1}, we can get the images of different intrinsic parameters by adjusting the resize rates. Motivated by DD3D~\cite{park2021pseudo}, we decouple depth from different focal lengths and acquire the scale-invariant depth as 

\begin{equation}
  d = \frac{s}{c} \cdot d_m,
  \label{eq:2}
\end{equation}

\begin{equation}
  s = \sqrt{\frac{1}{f_x^2} + \frac{1}{f_y^2}},
  \label{eq:3}
\end{equation}
where $d_m$ is the metric depth, $s$ is the original pixel size, and $c$ is a constant representing the pixel size at a given reference focal length. Through~\cref{eq:2}, 
the pixel scale of the reference focal length is regarded as the basis for depth estimation, which makes the predicted depth consistent with the size of the object in the image.
During the training, we modify the intrinsic parameters and image resolutions simultaneously. Once obtaining the scale-invariant depth, we utilize the actual focal length to encode the estimated depth to the metric depth, which greatly alleviates the problem caused by different intrinsic parameters among domains.

\subsubsection{Dynamic Perspective Augmentation}
\label{subsec:pose_aug}
Camera poses relative to the ego car are usually divergent among different domains. As noted in~\cite{zhao2021camera}, monocular depth predictors are naturally biased $w.r.t$ the distribution of camera poses, which inevitably impairs the accuracy of depth estimates when inferring on the unseen target domain. Transforming the image perspective in the source domain can be a feasible solution to obtain more robust depth predictions.
 However, direct perturbation of image perspective  (\ie camera pose) like PDA~\cite{zhao2021camera} is not feasible due to the unavailability of pixel-wise depth, instead, we propose dynamic perspective augmentation  by leveraging
 homography~\cite{dubrofsky2009homography} to heuristically generate various perspective images for model learning.

A homography is a mapping between two planar surfaces which is widely used for perspective conversion. The homography matrix $H \in \mathbb{R}^{3 \times 3}$ projects $p_1$ on one plane to $p_2$ on another plane, 

\begin{equation}
  sp_2 = Hp_1,
  \label{eq:important}
\end{equation}
where $p = [x, y, 1]^T$ is the homogeneous coordinate of a 2D point in a plane and $s$ is the scale factor. Since the homography matrix has natural attributes with 8 degrees of freedom, at least 4 corresponding point pairs are needed for recovering the matrix. 


Suppose camera pose relative to the ego car as $P_i = (y_i, p_i, r_i)$, where $i$ is the index of camera, $y_i$, $p_i$ and $r_i$ denote yaw, pitch and roll respectively. Then we perturb the camera pose as 
\begin{equation}
  \hat{P_i} = (y_i+\Delta y_i, p_i+\Delta p_i, r_i+\Delta r_i),
  \label{eq:important}
\end{equation}
where $\Delta y_i$, $\Delta p_i$ and $\Delta r_i$ are the random perturbation. We opt to use the homography matrix to describe the projection relationship between the original camera pose and the scrambled camera pose. 
Let $B_{gt} = \{b_1, \cdots, b_n\}$ represent the 3D ground truth boxes, where $n$ is the number of boxes. We pick up five bottom points $Q_{gt} = [x_{gt}, y_{gt}, z_{gt}]^T$ of 3D ground truth box $b_i$ as representatives, including one bottom center point and four bottom corner points. 
The selected points $Q_{gt}$ will be transformed into the original image plane by spatial mapping relationship, which is defined by

\begin{equation}
  d \cdot q = K  (\Phi(P) \cdot Q_{gt} + T),
  \label{eq:important}
\end{equation}
where $d$ is the actual depth, $\Phi$ is the transformation of Euler angles into a rotation matrix, $T$ is the transformation matrix from the ego car to the camera and $K$ denotes the intrinsic matrix. At the same time, the identical points $Q_{gt}$ will be transformed into the scrambled image plane by
\begin{equation}
  \hat{d} \cdot \hat{q} = K  (\Phi(\hat{P}) \cdot Q_{gt} + T).
  \label{eq:important}
\end{equation}
If $q$ and $\hat{q}$ are both in the range of image size, this pair of points will be kept. The formulation of the transformation of the two perspectives is defined as
\begin{equation}
  \hat{q} = Hq,
  \label{eq:important}
\end{equation}
When more than four pairs of points are reversed for a camera, we can acquire the estimated homography matrix $H$ by applying the least square method. Then we can utilize the homography matrix to roughly convert the original image into the one after the camera pose perturbation. More details about the homography principles and implementation can be found in the \emph{supplementary materials}.

\subsubsection{Domain-Invariant Feature Learning} 
\label{subsec:domain_inv_feat}
Domain-related annotations\cite{chen2018domain,
ganin2015unsupervised}, which are used to extract domain-agnostic representations containing intrinsic characteristics, is helpful for improving the generalization capability of models. However, when the target domain is inaccessible, how to well extract the domain-agnostic representations remains under-explored.

Intrinsic characteristics in the domain include image style, illumination, object scale, \etc And in MV3D-Det, one of the most significant differences among domains is the object scale caused by the diverse intrinsic parameters, which is also an important reason for the shift of feature distribution.
From \cref{sec:3.3.1}, we conclude that random scaling of images indicates a corresponding change in the intrinsic parameters. Hence, to acquire domain-invariant feature representation, we enforce the network to classify the domain itself by explicitly constructing pseudo-domain categories based on the focal length values.

Considering the wide range of the focal length values of camera intrinsic parameters, we quantize the focal length interval [$\alpha$, $\beta$] into $K$ sub-intervals by uniform discretization (UD), where $\alpha$ and $\beta$ denote the minimum and maximum values of the interval, respectively. The discretization thresholds $t_i \in \{t_0, t_1, \cdots , t_K\}$ can be formulated as:

\begin{equation}
  t_i = \alpha + \frac{(\beta - \alpha) * i}{K},
  \label{eq:important}
\end{equation}

\begin{figure}[h]
  \centering
   \includegraphics[width=0.8\linewidth]{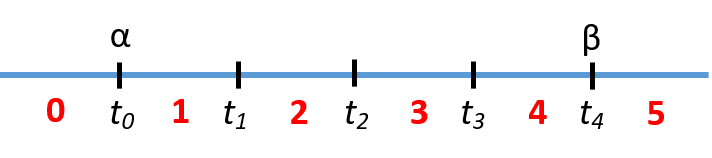}
   \vspace{-1em}
   \caption{Illustration of the relationships among sub-intervals, discrete focal distance values and categories. $t_i$ represents the discrete value, and $\{0, 1, 2, 3, 4, 5\}$ denote corresponding the domain categories. In this figure, there are 4 sub-intervals, 5 discretization thresholds, and 6 categories.}
   \label{fig:label_assign}
\end{figure}

Assuming that different focal length sub-intervals represent different pseudo-domains, it is obvious that the pseudo-domains form a well-ordered set with a strong ordinal correlation. However, typical classification losses (\eg CrossEntropy Loss, Focal Loss \cite{lin2017focal}) ignore the ordered information among the discrete labels. 
Motivated by~\cite{fu2018deep}, we treat the pseudo-domain classification as a sequential process and adopt an ordinal loss to make the most of the ignored information. Due to the intrinsic properties of ordinal classification, we opt to take ranges on both sides of the interval into consideration instead only closed intervals. As shown in \cref{fig:label_assign}, if the interval originally has $K$ sub-intervals, it means that there are $K+1$ discretization thresholds and $K+2$ categories. 

Let $x$ denote the feature map given an image, then  we can acquire the ordinal outputs by

\begin{equation}
  y = \phi (x, \theta),
\end{equation}
where $y$ is a 2$(K+1)$-dimensional vector and $\theta$ is the parameters of the domain classifier. Let $l \in \{0,1, \cdots, K+1\}$ denote the discrete label.
Our ordinal loss can be defined as 
\begin{gather}
  \mathcal{L}(y, l) = \sum_{k=0}^{K+1} \gamma(k, l) log(P^k) + (1-\gamma(k, l))log(1-P^k), \nonumber  \\
  \gamma(k, l) = 
  \begin{cases}
    1,& l \leq k \\
    0,& l > k 
  \end{cases}  \\
  P^k = \frac{e^{y_{(2k)}}}{e^{y_{(2k)}} + e^{y_{(2k+1)}}},  \nonumber
\end{gather}
where $\gamma(k, l)$ indicates whether the actual focal length value is less than the $k$-th discrete focal length threshold and $P^k$ denotes the probability that the domain classifier discriminates the focal length less than the $k$-th discrete values.

To align the domain distribution, we simultaneously optimize the domain classifier to minimize the ordinal loss and the base network to maximize this loss. For the implementation we use the gradient reverse layer (GRL)~\cite{ganin2015unsupervised} to invert the gradient back from the domain classifier.

\section{Experiments}
\subsection{Experimental Setup}
\noindent\textbf{Datasets.} We conduct experiments on three widely used autonomous driving datasets: nuScenes~\cite{caesar2020nuscenes}, Waymo~\cite{sun2020scalability}, and Lyft~\cite{lyft}. Each dataset has a diverse set of cameras with different intrinsic parameters and extrinsic parameters. We summarize the dataset information in detail in the \emph{supplementary material}.

\noindent\textbf{Comparision Methods.} In our experiments, we compare our DG-BEV with three counterparts: (\emph{i}) \textbf{Source Only} indicates directly using the model trained by the source domain to evaluate on the target domain. (\emph{ii}) \textbf{Oracle} indicates the fully supervised model trained on the target domain. (\emph{iii}) \textbf{CAM-Convs}~\cite{facil2019cam} is a new type of convolution that improves the generalization capabilities of depth prediction networks considerably. 

\noindent\textbf{Evaluation Metrics.} The mAP defined by nuScenes is based on the matching of 2D center distance on the ground plane instead of the Intersection over Union (IoU), which measures the error of ranging. Hence, we adopt the same validation metrics predefined officially by nuScenes for all datasets for simplicity. Since the attribute labels and the velocity labels are different from each other, we discard the Average Attribute Error (mAAE) and the Average Velocity Error (mAVE) in the case that nuScenes is not the source domain, and report the Average Precision (mAP), the Average Translation Error (mATE), the Average Scale Error (mASE) and the Average Orientation Error (mAOE) for other cases. Due to the lack of necessary metrics (\ie mAAE and mAVE), we develop NDS$^*$ as an alternative, which is defined as:

\begin{equation}
\setlength{\abovedisplayskip}{2pt}
\setlength{\belowdisplayskip}{3pt}
  \rm{NDS^* = \frac{1}{6}[3 \, mAP + \sum_{mTP \in \mathbb{TP}}(1-min(1,mTP))]},
  \label{eq:important}
\end{equation}
We focus on the commonly used vehicle category, and more specifically, the `car', `truck', `construction vehicle', `bus', and `trailer' of nuScenes, the `vehicle' of Waymo and the `car' of Lyft. What is more, to maintain consistency during training and validation, we only validate results in the range [-50m, 50m] like nuScenes.

\noindent\textbf{Implementation Details.} To validate the effectiveness of our DG-BEV, we adopt BEVDepth as our base model. Following~\cite{huang2021bevdet}, models are trained with AdamW~\cite{loshchilov2018fixing} optimizer, in which gradient clip is exploited with learning rate 2e-4, a total batch size of 64 on 8 Tesla V100s. We use $W_{in} \times H_{in}$ to denote the width and height of the input image and $W \times H$ represents the origin resolution. Then the original image will be processed by random flipping, random scaling with a range of $s \in [W_{in}/W-0.04, W_{in}/W+0.18]$ for nuScenes and Lyft and $s \in [W_{in}/W-0.08, W_{in}/W+0.08]$ for Waymo, 
random rotating with a range of $r \in [-5.4^\circ, 5.4^\circ]$ and finally cropping to a size of $W_{in} \times H_{in}$. Due to the different aspect ratios of different datasets, We use $704 \times 256$ as the input size for nuScenes, $704 \times 320$ for Waymo and $704 \times 384$ for Lyft. More implementation details are shown in the \emph{supplementary material}.


\subsection{Main Results}
 As shown in~\cref{tab:1}, we compare the detection performance with Source Only, Oracle, and CAM-Convs. Our method outperforms the Source Only baseline and CAM-Convs baseline under four different settings. We can observe that CAM-Convs hardly improve the performance of the model on the target domain. On nuScenes$\rightarrow$Waymo and Waymo$\rightarrow$nuScenes tasks, the Source Only model cannot detect 3D objects where the mAP almost drops to 0 caused by the huge domain gap. Our approach can greatly enhance the generalization ability of the model and achieves 64\% and 80\% of Oracle performance (NDS$^*$) in nuScenes$\rightarrow$Waymo and Waymo$\rightarrow$nuScenes, respectively. As for the nuScenes$\rightarrow$Lyft and Lyft$\rightarrow$nuScenes tasks, the Source Only model still maintains a certain level of detection capability, which indicates that there is no extremely large domain gap between the two. The reason is that the camera intrinsic parameters of the two datasets are close to each other and the six multi-view camera poses are similar. In this situation, our method can greatly improve the performance of the model in the unknown domain, \eg 0.296 NDS$^* \rightarrow$0.437 NDS$^*$ in nuScenes$\rightarrow$Lyft and 0.213 NDS$^* \rightarrow$0.374 NDS$^*$ in Lyft$\rightarrow$nuScenes.

\begin{table*}
  \centering
  \begin{tabular}{c|ccccc|ccccc}
    \toprule
    Nus $\rightarrow$ \rm{Waymo} & \multicolumn{5}{c|}{Source Domain (nuScenes)} &\multicolumn{5}{c}{Target Domain (Waymo)} \\
    \midrule
    Method & mAP$\uparrow$ & mATE$\downarrow$ & mASE$\downarrow$ & mAOE$\downarrow$ & NDS$\uparrow$ & mAP$\uparrow$ & mATE$\downarrow$ & mASE$\downarrow$ & mAOE$\downarrow$ & NDS\^*$\uparrow$\\
    \midrule
    Oracle &-&-&-&-&-&0.552&0.528&0.148&0.085&0.649\\
    \midrule
    Source Only &0.328 &0.666 &0.274 &\textbf{0.560} &0.407 &0.040 &1.303 &0.265 &0.790 &0.178 \\
    CAM-Convs~\cite{facil2019cam} &0.328 &0.681 &0.273 &0.571 &0.397 &0.045 &1.301 &0.253 &0.773 &0.185\\
    DG-BEV (Ours) &\textbf{0.337}&\textbf{0.647}&\textbf{0.272}&0.567&\textbf{0.407}& \textbf{0.297} &\textbf{0.822}& \textbf{0.216} &\textbf{0.372}& \textbf{0.415}\\

    \toprule
    Waymo $\rightarrow$ Nus & \multicolumn{5}{c|}{Source Domain (Waymo)} &\multicolumn{5}{c}{Target Domain (nuScenes)} \\
    \midrule
    Method & mAP$\uparrow$ & mATE$\downarrow$ & mASE$\downarrow$ & mAOE$\downarrow$ & NDS\^*$\uparrow$ & mAP$\uparrow$ & mATE$\downarrow$ & mASE$\downarrow$ & mAOE$\downarrow$& NDS\^*$\uparrow$\\
    \midrule
    Oracle &- &- &- &- &- &0.475 &0.577 &0.177 &0.147 &0.587 \\
    \midrule
    Source Only &0.552&0.528&\textbf{0.148}&0.085&0.649&0.032 &1.305 &0.768 & 0.532 &0.133\\
    CAM-Convs~\cite{facil2019cam} &0.549 &0.532&0.148 &0.080 &0.648 &0.038 &1.308 &0.316 &0.506 &0.215\\
    DG-BEV (Ours) &\textbf{0.568}&\textbf{0.519}&0.149&\textbf{0.078}&\textbf{0.660}&\textbf{0.303}&\textbf{0.689}&\textbf{0.218}&\textbf{0.171}&\textbf{0.472}\\

    \toprule
    Nus $\rightarrow$ \rm{Lyft} & \multicolumn{5}{c|}{Source Domain (nuScenes)} &\multicolumn{5}{c}{Target Domain (Lyft)} \\
    \midrule
    Method & mAP$\uparrow$ & mATE$\downarrow$ & mASE$\downarrow$ & mAOE$\downarrow$ & NDS$\uparrow$ & mAP$\uparrow$ & mATE$\downarrow$ & mASE$\downarrow$ & mAOE$\downarrow$& NDS\^*$\uparrow$\\
    \midrule
    Oracle &-&-&-&-&-&0.602 &0.471 &0.152 &0.078 &0.684\\
    \midrule
    Source Only &0.328 &0.666 &0.274 &0.560 &0.407
    &0.112 &0.997 &0.176 &0.389 &0.296\\
    CAM-Convs~\cite{facil2019cam} &0.328 &0.681 &0.273 &0.571 &0.397 &0.145 &0.999 &0.173 &0.368 &0.316\\
    DG-BEV (Ours) &\textbf{0.341}&\textbf{0.655}&\textbf{0.273}&\textbf{0.538}&\textbf{0.409}&\textbf{0.287}&\textbf{0.771}&\textbf{0.170}&\textbf{0.302}&\textbf{0.437}\\

    \toprule
    Lyft $\rightarrow$ Nus & \multicolumn{5}{c|}{Source Domain (Lyft)} &\multicolumn{5}{c}{Target Domain (nuScenes)} \\
    \midrule
    Method & mAP$\uparrow$ & mATE$\downarrow$ & mASE$\downarrow$ & mAOE$\downarrow$ & NDS\^*$\uparrow$ & mAP$\uparrow$ & mATE$\downarrow$ & mASE$\downarrow$ & mAOE$\downarrow$& NDS\^*$\uparrow$\\
    \midrule
    Oracle&-&-&-&-&-&0.401 &0.651 &0.179 &0.484 &0.482\\
    \midrule
    Source Only &0.602 &0.471 &0.152 &0.078 &0.684 &0.102 &1.143 &0.239 &0.789 &0.213 \\
    CAM-Convs~\cite{facil2019cam} &\textbf{0.611}&\textbf{0.465}&\textbf{0.149}&\textbf{0.075}&\textbf{0.691}&0.098 &1.198 &0.209 &1.064 &0.181\\
    DG-BEV (Ours) &0.590&0.488&0.153&0.079&0.675&\textbf{0.268} &\textbf{0.764} &\textbf{0.205} &\textbf{0.591} &\textbf{0.374}\\
    \bottomrule
  \end{tabular}
  \caption{Performance of DG-BEV on four source-target pairs. For the source domain, we report the average metric for the 10 categories of nuScenes. For the target domain, we report the average metric for the car category when Lyft$\rightarrow$nuScenes and the metric of five categories when Waymo$\rightarrow$nuScenes, including car, truck, construction vehicle, bus and trailer.
  As for Lyft and Waymo, we report the results of the vehicle category and car category, respectively. 
  All results are on the validation subset of the corresponding dataset. 
  }
  \label{tab:1}
\end{table*}

\subsection{Ablation Studies and Analysis}
In this section, we explore the role of each module in DG-BEV through more detailed ablation studies. If not specified, all experiments are conducted with BEVDepth-R50 on the task of training on nuScenes and validating on 1/2 subset of Waymo.


\subsubsection{Main Ablations}

In order to understand of how each component contributes to the final performance, we subsequently add the proposed module and report the performance in \cref{tab:main_ablations}. The vanilla baseline starts from 0.178 NDS$^*$, which incurs a drastic performance drop compared to the source domain. When intrinsic-decoupled depth estimation module is added, the detection accuracy improves from 0.178 to 0.393, indicating the necessity of the disentanglement between depth estimation and camera intrinsic. Then, we apply the dynamic perspective augmentation strategy, which further gains 1.4\% NDS$^*$. Finally, when domain-invariant feature learning is introduced, the performance achieves 0.415 NDS$^*$, yielding an enhancement of 24\% NDS$^*$.

\begin{table}[!h]
    \centering
    \begin{tabular}{ccc|cc}
        \toprule
         IDD & DPA & DIFL &  mAP $\uparrow$ & NDS\^* $\uparrow$\\
         \midrule
           & &  & 0.040 & 0.178\\
          $\checkmark$ & & & 0.272 & 0.393\\
          $\checkmark$ & $\checkmark$ & & 0.297 & 0.408\\
          $\checkmark$ & $\checkmark$ & $\checkmark$ & 0.297 & 0.415\\
        \bottomrule
    \end{tabular}
    \caption{Ablation studies on the effectiveness of each component in DG-BEV. ``IDD'' denotes intrinsic-decoupled depth estimation, ``DPA'' denotes dynamic perspective augmentation, and ``DIFL'' denotes domain-invariant feature learning.}
    \label{tab:main_ablations}
\end{table}

\subsubsection{Dynamic Perpsective Augmentation} 
In order to investigate the effect of camera pose perturbation on the generalization ability of the model, we conducted experiments under two settings, nuScenes$\rightarrow$Waymo (6 cameras$\rightarrow$5 cameras) and nuScenes$\rightarrow$Lyft (6 camera$\rightarrow$6 cameras), respectively. The results are shown in \cref{tab:3}. Overall, perspective augmentation can improve the generalization capability of the model in both settings. 
Specifically, for nuScenes$\rightarrow$Waymo, perturbation of pitch within a certain range can lead to improved results (0.284 mAP), and perturbation of yaw always brings a definite gain. The reason is that the difference in extrinsic parameters between nuScenes and Waymo is primarily due to different yaws of the camera relative to the ego car. For nuScenes$\rightarrow$Lyft, a perturbation of pitch can promote the detection ability greatly, where the possible reason is that the heights of the cameras are different when collecting the two datasets, so adjusting the pitch can mitigate the difference to some extent. Also since nuScenes and Lyft are both multi-view 6 cameras and each camera has a similar orientation, perturbing the yaw only obtains a little improvement. Moreover, since roads are not always flat, adjusting the roll range allows the model to adapt to different slopes, and enhance the generalization capability of the model in both settings.

\begin{table}
  \centering
  \begin{tabular}{c|c|c}
    \toprule
     & nuScenes $\rightarrow$ Waymo &nuScenes $\rightarrow$ Lyft \\
    Perturbation & mAP$\uparrow$ &mAP$\uparrow$  \\
    \midrule
    0 & 0.272 & 0.242 \\
    \midrule
    $\Delta$ p=0.01 & 0.284 & 0.260 \\
    $\Delta$ p=0.02 &0.259 & \textbf{0.293} \\
    $\Delta$ p=0.03 &0.258 & 0.270  \\
    $\Delta$ p=0.04 & 0.247& 0.275  \\ 
    \midrule
    $\Delta$ y=0.02 &0.286 & 0.252 \\ 
    $\Delta$ y=0.04 &0.281 & 0.248 \\
    $\Delta$ y=0.06 & 0.277 & 0.254\\
    $\Delta$ y=0.08 & 0.283 &0.252\\
    \midrule
    $\Delta$ r=0.02 & \textbf{0.290} &0.263 \\
    $\Delta$ r=0.04 & 0.286 &0.263 \\
    $\Delta$ r=0.06 & 0.283 &0.270\\
    $\Delta$ r=0.08 & 0.286 &0.263\\
    \bottomrule
  \end{tabular}
  \caption{Ablation study of Dynamic Perspective Augmentation. $\Delta$ p, $\Delta$ y and $\Delta$ r denote the perturbed range of pitch, yaw and roll, respectively.}
  \label{tab:3}
\end{table}


\subsubsection{Domain-Invariant Feature Learning} 
The domain shift among different domains contains many factors, and we think that one of the most important factors in 3D detection is scale variation caused by the focal length. To verify the effectiveness of our method, we adopt the classical method of DA~\cite{ganin2015unsupervised} including both source and target domain and conduct detailed experiments in \cref{tab:4}. No matter whether the target domain is included or not, using domain classifiers to align feature distributions among different domains does not bring any gains (0.171 NDS$^*$ \& 0.002 NDS$^*$). This indicates that the domain shift across domains can not be resolved only from the feature dimension. Moreover, under the setting of decoupling depth, our proposed method can achieve the improvement of 1.4 NDS$^*$ (0.393 $\rightarrow$ 0.407), while the result has a serious degradation (0.198 NDS$^*$) after the target domain is available. The reasons  can be two-fold: (i) directly aligning the feature distributions between the source and target domain would make the network focus on all inter-domain differences without distinction, while most of the inter-domain differences may not be conducive to accurately estimating the spatial information of the objects. (ii) the images of different domains have different aspect ratios. When both are input to the network with the same size, it is necessary to pad the image of one of the domains, which makes it easy for the domain classifier to identify domains based on whether images are padding. As a result, the image features are filled with noise after passing the GRL, which causes the degradation of the results.


\noindent\textbf{Domain classification loss.} Domain classifier with GRL usually performs a classification task to align the feature distributions between the source and target domain. In this paper, we divide the input images into different domains according to the ranges of the focal lengths. Since the focal length has the property of order, we explore the effect of different classification losses on performance. The experiment is based on the premise of intristic-decoupled depth, and the results are shown in \cref{tab:5}. We can find that the CrossEntropy loss and the focal loss can not work well (0.397 NDS$^*$ \& 0.401 NDS$^*$) in such a ordinal classification. In contrast, with the help of the ordinal loss, our model reaches 0.407 NDS$^*$.

\begin{table}
  \centering
  \begin{tabular}{ccc|cccc}
    \toprule
    DC  & IDD & Target Domain & mAP$\uparrow$ & NDS\^*$\uparrow$ \\
    \midrule
    &&&0.040 &0.178  \\
    \checkmark&&&0.032 &0.171  \\
    \checkmark & &\checkmark&0.004 &0.002  \\
    &\checkmark&&0.272 &0.393  \\
    \checkmark&\checkmark&& \textbf{0.292} &\textbf{0.407} \\
    \checkmark &\checkmark &\checkmark &0.054 &0.198 \\
    \bottomrule
  \end{tabular}
  \caption{Ablation study of Domain-Invariant Feature Learning. ``DC'' denotes the domain classifier and ``IDD'' denotes intrisic-decoupling depth prediction.}
  \label{tab:4}
\end{table}

\begin{table}
  \centering
  \begin{tabular}{c|cccc}
    \toprule
    Loss & mAP$\uparrow$ & NDS\^*$\uparrow$  \\
    \midrule
    Cross Entropy Loss&0.282 &0.397  \\
    Focal Loss & 0.281 &0.401 \\
    Oridinal Loss&\textbf{0.292} &\textbf{0.407}    \\
    \bottomrule
  \end{tabular}
  \caption{Ablation study on three different domain classification loss.}
  \label{tab:5}
\end{table}

\section{Conclusion}
In this paper, we have proposed a novel domain-general BEV perception method named DG-BEV which can alleviate the performance
drop on the unseen target domain. We observe that current BEV perception methods are all for specific domain, which will greatly limit the application in the industry. We decouple the BEV feature distribution with specific domain by the proposed instrinsics-decoupled depth prediction and domain-invariant feature learning. Extensive experiments on various public datasets, including Waymo, nuScenes, and Lyft, demonstrate the generalization and effectiveness of our approach. We hope that the proposed method DG-BEV could improve the implementation of BEV perception in the industry.

{\small
\bibliographystyle{ieee_fullname}
\bibliography{egbib}
}

\end{document}


\title{Towards Domain Generalization for Multi-view 3D Object Detection \\ in Bird-Eye-View Supplementary Material}

\author{First Author\\
Institution1\\
Institution1 address\\
{\tt\small firstauthor@i1.org}
\and
Second Author\\
Institution2\\
First line of institution2 address\\
{\tt\small secondauthor@i2.org}
}
\maketitle

\section*{Appendix}
\subsection*{A. Dataset Comparisons}
\begin{figure}[!t]
  \centering
   \includegraphics[width=1.0\linewidth]{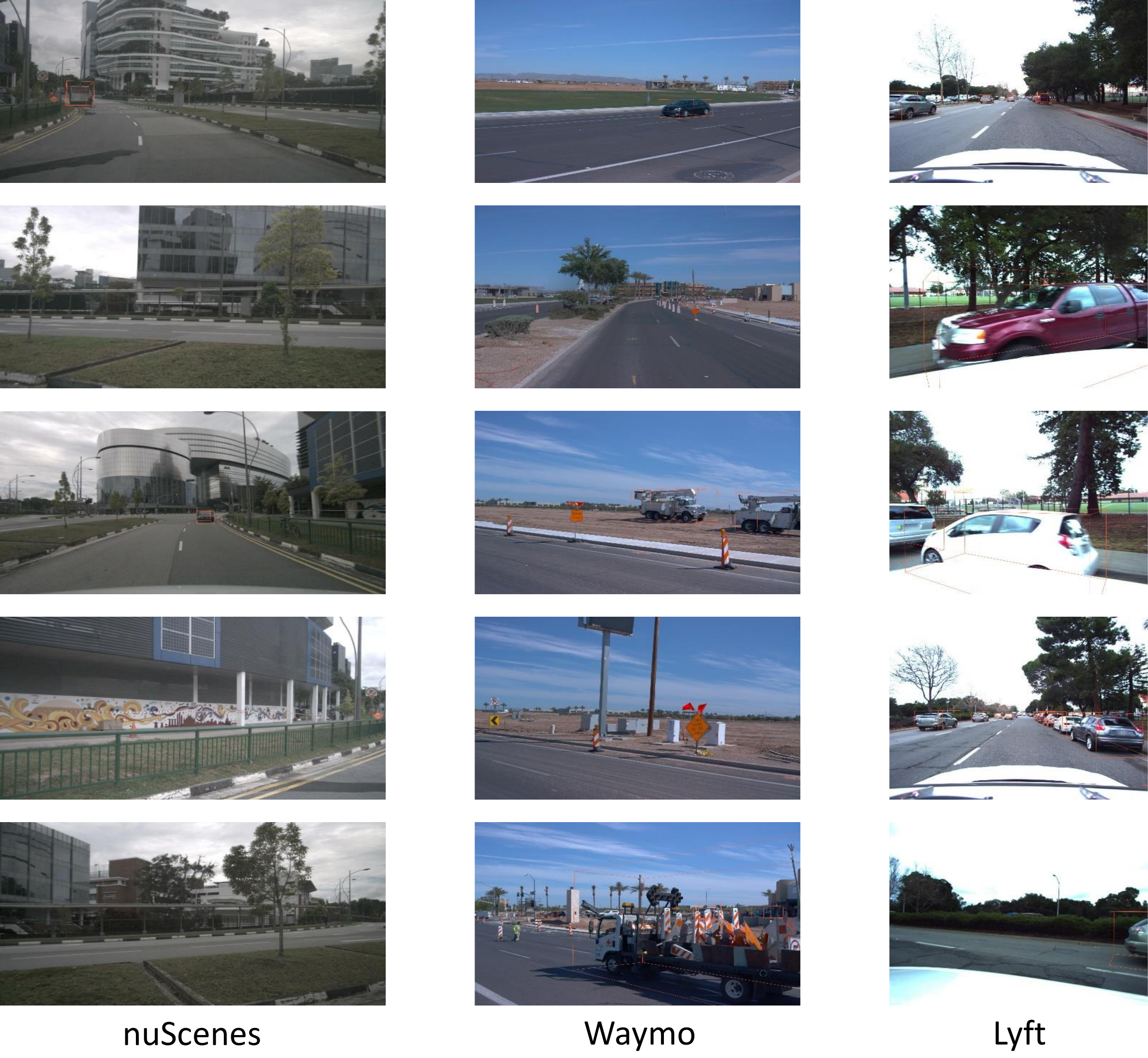}
   \vspace{-1em}
   \caption{Dataset visualizations with ground-truth labels.
}
   \label{fig:network}
\end{figure}

We summarize the dataset information of nuScenes~\cite{caesar2020nuscenes}, Waymo~\cite{sun2020scalability} and Lyft~\cite{lyft} in detail in~\cref{tab:0}. To provide more intuitive comparisons among different datasets, we present images with projected ground-truth labels in~\cref{fig:network}.
It is obvious that cameras utilized in these datasets are different, which is reflected in the image resolutions, object scale, \etc.

\begin{table*}[h]
  \centering
  \begin{tabular}{cccccccc}
    \midrule
    Dataset & Size  & Location & Shape & Number of Cameras & 360$^{\circ}$ & Object &Night \\
    \midrule
    nuScenes~\cite{caesar2020nuscenes}& 28130 &Boston,SG. &(900,1600) &6 &Yes &23 &Yes \\
    Waymo~\cite{sun2020scalability} &158081 &USA &(1280, 1920),(886, 1920) & 5& No &4 &Yes \\
    Lyft~\cite{lyft}& 18900 &Palo Alto &(1024, 1224) &6 &Yes &9 &No \\
    \midrule
  \end{tabular}
  \caption{Dataset Overview. The size refers to the number of images used in the training stage and 360$^{\circ}$ indicates whether the cameras cover a 360$^{\circ}$ view. SG: Singapore.}
  \label{tab:0}
\end{table*}

\subsection*{B. Dynamic Perspective Augmentation}
\label{sec:intro}

Suppose that two cameras shoot the plane $P$ at different poses. As shown in the figure, the normal vector of the plane in frame 1 is $N$ and the distance from $P$ to frame 1 is $d$. For the point $X_1$ in Frame 1, we can get

\begin{equation}
  N^T X_1 = d,
  \label{eq:1}
\end{equation}
At the same time, it can be converted to frame 2 by the following equation
\begin{equation}
  X_2 = RX_1 + T,
  \label{eq:2}
\end{equation}
By using~\cref{eq:1,eq:2}, we can get that
\begin{equation}
  X_2 = RX_1 + T \frac{1}{d} N^T X_1 \\
        = (R + T \frac{1}{d} N^T) X_1.
  \label{eq:3}
\end{equation}
Then the 3D coordinate can be projected to the 2D image plane as:
\begin{equation}
  x_1 = \frac{1}{z_1}KX_1, \\
  \label{eq:4}
\end{equation}
where $K$ is the intrinsic parameters of the camera and $z$ is the depth.

By~\cref{eq:3,eq:4}, we can get that
\begin{equation}
  x_2  = \frac{z_2}{z_1} K^{-1} (R + T \frac{1}{d} N^T) K x_1 \\
  = \frac{z_2}{z_1} H x_1,
  \label{eq:5}
\end{equation}
where $x_1$ and $x_2$ are the homogeneous coordinates. Then we can get the homography matrix:
\begin{equation}
 H  =  K^{-1} (R + T \frac{1}{d} N^T) K, \qquad H \in \mathbb{R}^{3 \times 3}.
  \label{eq:6}
\end{equation}
Since $\frac{z_2}{z_1}$ does not affect the validity of~\cref{eq:5}, we can determine one element in $H$ to be 1 and the homography matrix has 8 degrees of freedom. So at least 4 corresponding point pairs are needed for recovering the matrix.

When the camera is only spinning ($T=0$) or moving a small distance ($T \rightarrow 0$), the homography matrix is independent of $d$ and $N$, which indicates~\cref{eq:5} can be applied to the entire 3D space, rather than a plane. Therefore, we can leverage homography~\cite{dubrofsky2009homography} to heuristically generate various perspective images for model learning. The implementation details are shown in~\cref{alg1}.

\begin{algorithm} [!t]
    \caption{Dynamic Perspective Augmentation}
    \label{alg1}
    \begin{algorithmic}[1]
       \REQUIRE Multi-View Images \textbf{I} = $\{ I_1, I_2, \cdots , I_N\}$, Rotation Matrix from Ego Car to Camera \textbf{R} = $\{ R_1, R_2, \cdots , R_N\}$, Translation Matrix from ego car to Camera \textbf{T} = $\{ T_1, T_2, \cdots , T_N\}$, Intrinsic Parameters of Cameras \textbf{K} = $\{ K_1, K_2, \cdots , K_N\}$, $N$ is the number of Surround Cameras, the Transformation of Rotation Matrix into Euler Angles $\phi$, the Transformation of Euler Angles into Rotation Matrix $\Phi$, Bottom Center and Bottom Corner Set of Ground Truth \textbf{O} = $\{ O_1, O_2, \cdots , O_M\}$, Yaw Range $[y_{min}, y_{max}]$, Pitch Range $[p_{min}, p_{max}]$, Roll Range $[r_{min}, r_{max}]$, Paired Matching Points Set $L$.
    
    \FOR{$n$ in range$(0, N)$} 
       \FORALL{$m$ such that $O_m = (x_m, y_m, z_m) \in$ \textbf{O}}
        \STATE $d \cdot (u, v, 1)^T = K_n(R_n \cdot O_m + T_n)$;
        \STATE // whether $O_m$ is projected to the current image
        \IF{$d > 0$ and $(u, v)$ in $I_n.size()$} 
           \STATE $\Delta y = SAMPLE(y_{min}, y_{max})$; 
           \STATE $\Delta p = SAMPLE(p_{min}, p_{max})$;
           \STATE $\Delta r = SAMPLE(r_{min}, r_{max})$;
            \STATE // get the origin camera pose
           \STATE $yaw, pitch, roll = \phi(R_n)$;
            \STATE // perturb the camera pose
           \STATE $R_n^{\prime} = \Phi(yaw+\Delta y, pitch+\Delta p, roll+\Delta r)$;
           \STATE $d^{\prime} \cdot (u^{\prime}, v^{\prime}, 1)^T = K_n(R_n^{\prime} \cdot O_m + T_n)$;
           \STATE // whether $O_m$ is projected to perturbed image
           \IF{$d^{\prime} > 0$ and $(u^{\prime}, v^{\prime})$ in $I_n.size()$} 
              \STATE $L.append([(u, v), (u^{\prime}, v^{\prime})])$;
           \ENDIF
        \ENDIF
        
    \ENDFOR
    \STATE // Whether over four pairs of points are matched
   \IF{$len(L) >= 4$} 
        \STATE // $LS$ denotes least squares method 
        \STATE  $H = LS(L)$;
        \STATE $I^{\prime}_n = H \cdot I_n$;
   \ELSE
        \STATE $I^{\prime}_n = I_n$;
   \ENDIF
   \ENDFOR

    \renewcommand{\algorithmicensure}{\textbf{Output:}}
    \ENSURE  Multi-View Perturbed Images $\textbf{I}^{\prime} = \{ I^{\prime}_1,  \cdots , I^{\prime}_N\}$
    \end{algorithmic}
\end{algorithm}

\subsection*{C. Detailed Training Settings}
In this section, we introduce more detailed training settings. As for the model, we follow the basic config provided in~\cite{huang2021bevdet}. Scale-invariant depth is determined by the metric depth and intrinsic parameters, so we employ [2, 90] for nuScenes, [1, 60] for Waymo and [1, 90] for Lyft. For the construction of the pseudo-domain categories, the settings of discretization thresholds are [500, 550, 600, 650, 700, 750] for nuScenes, [600, 650, 700, 750, 800, 850, 900] for Waymo and [500, 550, 600, 650] for Lyft.

\subsection*{D. Additional Experimental Results}
Here, we provide some additional empirical results in the task of nuScenes$\rightarrow$Waymo, building on top of BEVDet~\cite{huang2021bevdet}. The results are shown in~\cref{tab:1}. The Source Only model cannot detect 3D objects where the mAP almost drops to 0 caused by the huge domain gap, just the same as BEVDepth~\cite{li2022bevdepth}. We observe that CAM-Convs~\cite{facil2019cam} hardly improves the performance of the model on the target domain. By contrast, our approach greatly enhances the generalization ability of the model and achieves 64\% NDS$^*$ of Oracle performance, which verifies the generalization of DG-BEV.

\begin{table*}[h]
  \centering
  \begin{tabular}{c|ccccc|ccccc}
    \toprule
    Nus $\rightarrow$ \rm{Waymo} & \multicolumn{5}{c|}{Source Domain (nuScenes)} &\multicolumn{5}{c}{Target Domain (Waymo)} \\
    \midrule
    Method & mAP$\uparrow$ & mATE$\downarrow$ & mASE$\downarrow$ & mAOE$\downarrow$ & NDS^*$\uparrow$ & mAP$\uparrow$ & mATE$\downarrow$ & mASE$\downarrow$ & mAOE$\downarrow$ & NDS^*$\uparrow$\\
    \midrule
    Oracle &-&-&-&-&-&0.487&0.582&0.147&0.078&0.609\\
    \midrule
    Source Only &0.476 &0.587 &0.178 &0.143 &0.587 &0.028 &1.354 &0.273 &0.738 &0.179 \\
    CAM-Convs~\cite{facil2019cam} &0.477 &0.580 &0.177 &0.136 &0.590 &0.034 &1.346 &0.273 &0.721 &0.185\\
    DG-BEV (Ours) &\textbf{0.489}&\textbf{0.573}&\textbf{0.174}&\textbf{0.131}&\textbf{0.598}& \textbf{0.338} &\textbf{0.789}& \textbf{0.202} &\textbf{0.267}& \textbf{0.459} \\
    \bottomrule
  \end{tabular}
  \caption{Performance of DG-BEV in nuScenes$\rightarrow$Waymo task on top of BEVDet.
  }
  \label{tab:1}
\end{table*}

\subsection*{E. Broader Impacts}
In autonomous driving, updating of cameras and iterations of vehicles often occur, which lead to a sharp decline in the performance of the previously trained model in real-world scenarios. In this paper, we use different domains to represent the above existing problems and propose DG-BEV, a domain generalization method for multi-view 3D object detection, which successfully alleviates the performance drop on the unseen target domain without impairing the accuracy of the source domain. 

As for the limitation of our method, we do not take into account all the differences between domains and there are still numerous unsolved issues such as different color styles, various distributions of object dimensions, \etc
We hope that our proposed DG-BEV can be a well-developed baseline for future research.

{\small
\bibliographystyle{ieee_fullname}
\bibliography{egbib}
}